\documentclass[a4paper,twoside]{article}

\usepackage{epsfig}
\usepackage{subcaption}
\usepackage{calc}
\usepackage{amssymb}
\usepackage{amstext}
\usepackage{amsmath}
\usepackage{amsthm}
\usepackage{multicol}
\usepackage{pslatex}
\usepackage{apalike}
\usepackage{algorithm2e}
\usepackage[bottom]{footmisc}
\usepackage{booktabs}

\usepackage{tikz}
\usetikzlibrary{arrows}
\usetikzlibrary{arrows.meta}

\definecolor{aliceblue}{rgb}{0.94, 0.97, 1.0}
\definecolor{lavenderblush}{rgb}{1.0, 0.94, 0.96}
\definecolor{honeydew}{rgb}{0.94, 1.0, 0.94}
\definecolor{bananamania}{rgb}{0.99, 0.93, 0.78} 
\definecolor{magnolia}{rgb}{0.96, 0.95, 1.0}

\usepackage{SCITEPRESS}     

\usepackage[pagebackref=true,breaklinks=true,colorlinks,bookmarks=false]{hyperref}

\begin{document}

\title{Dance Style Classification using Laban-Inspired and Frequency-Domain Motion Features}

\author{\authorname{Ben Hamscher*, Arnold Brosch*, Nicolas Binninger, Maksymilian Jan Dejna and Kira Maag} 
\affiliation{Heinrich-Heine-University Düsseldorf, Germany}
\email{\{ben.hamscher, arnold.brosch, nicolas.binninger, maksymilian.dejna, kira.maag\}@hhu.de}
}


\abstract{
Dance is an essential component of human culture and serves as a tool for conveying emotions and telling stories. Identifying and distinguishing dance genres based on motion data is a complex problem in human activity recognition, as many styles share similar poses, gestures, and temporal motion patterns. This work presents a lightweight framework for classifying dance styles that determines motion characteristics based on pose estimates extracted from videos. We propose temporal-spatial descriptors inspired by Laban Movement Analysis. These features capture local joint dynamics such as velocity, acceleration, and angular movement of the upper body, enabling a structured representation of spatial coordination. To further encode rhythmic and periodic aspects of movement, we integrate Fast Fourier Transform features that characterize movement patterns in the frequency domain. The proposed approach achieves robust classification of different dance styles with low computational effort, as complex model architectures are not required, and shows that interpretable motion representations can effectively capture stylistic nuances.
}

\onecolumn \maketitle \normalsize \setcounter{footnote}{0} \vfill

\let\thefootnote\relax\footnotetext{* equal contribution}

%
%
%
\section{\uppercase{Introduction}}
\label{sec:introduction}
\begin{figure}[t]
  \centering
    \hfill
    \begin{subfigure}[t]{0.14\textwidth}
        \centering
        \includegraphics[width=\linewidth]{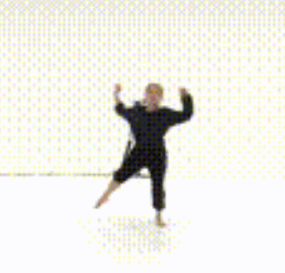}
        \caption*{Ballett Jazz} 
    \end{subfigure}
    \hfill
    \begin{subfigure}[t]{0.14\textwidth}
        \centering
        \includegraphics[width=\linewidth]{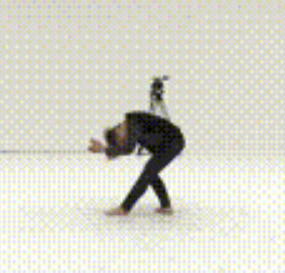}
        \caption*{Street Jazz} 
    \end{subfigure}
    \hfill
    \begin{subfigure}[t]{0.14\textwidth}
        \centering
        \includegraphics[width=\linewidth]{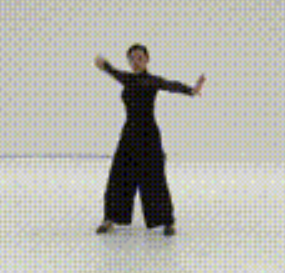}
        \caption*{Waack} 
    \end{subfigure}
    \hfill

    \vspace{1.5ex}
    \hfill
    \begin{subfigure}[t]{0.14\textwidth}
        \centering
        \includegraphics[width=\linewidth]{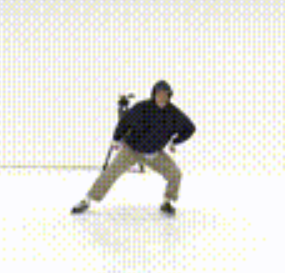}
        \caption*{Middle Hip-hop} 
    \end{subfigure}
    \hfill
    \begin{subfigure}[t]{0.14\textwidth}
        \centering
        \includegraphics[width=\linewidth]{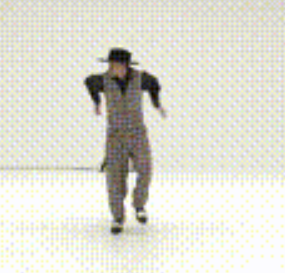}
        \caption*{Lock} 
    \end{subfigure}
    \hfill
    \begin{subfigure}[t]{0.14\textwidth}
        \centering
        \includegraphics[width=\linewidth]{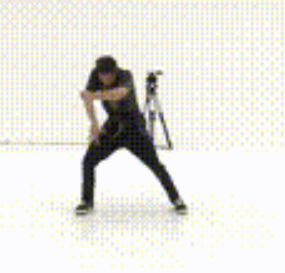}
        \caption*{Krump} 
    \end{subfigure}
    \hfill
    
    \vspace{1.5ex}
    \hfill
    \begin{subfigure}[t]{0.14\textwidth}
        \centering
        \includegraphics[width=\linewidth]{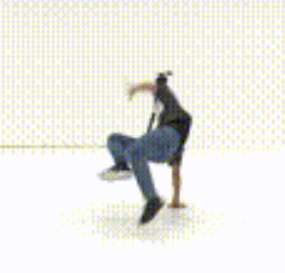}
        \caption*{Break} 
    \end{subfigure}
    \hfill
    \begin{subfigure}[t]{0.14\textwidth}
        \centering
        \includegraphics[width=\linewidth]{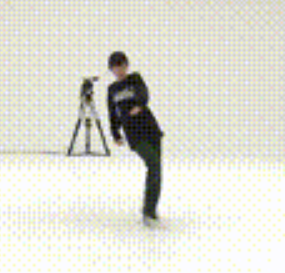}
        \caption*{Pop} 
    \end{subfigure}
    \hfill
    \begin{subfigure}[t]{0.14\textwidth}
        \centering
        \includegraphics[width=\linewidth]{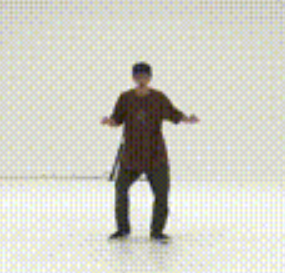}
        \caption*{House} 
    \end{subfigure}
    \hfill
  \caption{Sample images of different dance styles from the AIST dataset \cite{Tsuchida2019_aist}.}
  \label{fig:aist_examples}
\end{figure}

Dance has been a central part of human culture and expression for centuries. In almost all societies, it serves as a tool for conveying emotions, stories, and a sense of community. 
Different dance styles and forms have developed over time, reflecting the diversity of cultural identities as well as social and artistic processes of change. 
In addition to its aesthetic dimension, dance also fulfills a social function, i.e., it strengthens the sense of belonging within groups and acts as a universal form of communication that connects people regardless of language or origin. 
As a result of its rapid development, identifying and recognizing different dance styles becomes increasingly difficult and often requires expertise to distinguish them by visual observation. 
Classifying different dance styles is a particular challenge, as many dances feature similar postures, gestures, and movement sequences. Such subtle differences not only make precise classification difficult, but also impair the performance of the systems used. Example images for different dance genres are shown in Figure~\ref{fig:aist_examples}.

Dance-motion genre classification, which aims to recognize different dance styles, is a subfield of human activity recognition \cite{Jobanputra2019}. 
While human activity recognition typically focuses on everyday activities such as walking, running, sitting, or sleeping, the dance classification encompasses far more complex and subtle movements. 
Dances often involve rapid transitions, overlapping gestures, and stylistic variations that depend on rhythm, emotion, and cultural context, making them significantly more challenging to model and distinguish \cite{Gupta2024}. 
As a result, there is a growing demand for robust analytical tools that can help to identify and classify dance styles without relying solely on the knowledge of experts. 
Advances in machine learning, particularly in deep neural networks, offer excellent opportunities for automating this process by learning complex movement patterns directly from dance data. 
These approaches also include human pose estimation, which extracts skeletal representations from videos \cite{Zheng2023}. This enables the extraction and analysis of human movements, thus providing a more structured and interpretable understanding of dance movements and forming the basis for data-driven dance analysis and classification. 

A general framework for movement analysis is Laban Movement Analysis (LMA, \cite{Dewan2018_lma}), which is a comprehensive framework for observing, describing, and interpreting human body movement. 
It offers a structured method for analyzing movement based on four main components: body, effort, shape, and space. 
The category of \emph{body} focuses on which parts of the body are active and how they are coordinated, while \emph{effort} captures the qualitative dynamics of movement, such as weight, time, flow, and space. 
The category \emph{shape} describes how the body form changes during movement, and the category \emph{space} deals with spatial orientation and movement trajectories. 
LMA is often used in the fields of dance, performance studies, and movement therapy, as it bridges the expressive and physical aspects of movement \cite{Perez2025}.

In this work, we present a lightweight approach for dance style classification that incorporates features based on LMA and the Fast Fourier Transform (FFT, \cite{Brigham1967}). 2D as well as 3D body keypoints are extracted from the dance videos using a pose estimation model or collected by motion capture devices, and we then derive a set of motion-based features from them. 
Specifically, we calculate LMA-inspired features by analyzing joint movements relative to the hip center, which allows us to capture spatial characteristics of the dance. Note that this eliminates the need for normalization, as we consider all features relative to the center of the hip. 
From these relative joint positions, we calculate velocity and acceleration to quantify the dynamics of the movement. 
We also extract angular features to measure rotational movements of the upper body, as well as extremity distances, for example, of hands and feet, to describe the shape of the body. 
In addition, we calculate FFT-based features from the temporal joint trajectories to identify repetitive movement patterns, dominant frequencies, and rhythmic structures that are characteristic of different dance styles. 
The FFT breaks down the motion signal into its frequency components, showing which sine and cosine oscillations comprise the signal. 
Finally, these features are fed into lightweight machine learning models, such as regression, gradient boosting, or shallow neural networks, which learn to distinguish between dance genres.
In our experiments, we use various datasets such as AIST \cite{Tsuchida2019_aist}, Motorica Dance Dataset \cite{Alexanderson2023,Perez2022} and ImperialDance-Dataset \cite{Zhong2024}, and compare the performance of our features obtained from 2D or 3D keypoints, as well as basic and advanced dances.
The source code of our method is made publicly available at 
\url{https://github.com/benhamscher/dance-style-classification}.

We summarize our contributions as follows:
\begin{itemize}
    \item We construct features based on LMA and FFT, which distinguish the characteristics of different dance styles.
    \item We perform analyses using feature computation based on 2D and 3D keypoints, and we use various classification models.
    \item We conduct experiments on many different datasets, which contain up to $10$ different dance genres.
\end{itemize}
%
%
%
\section{\uppercase{Related Work}}
\label{sec:rel_work}

\paragraph{Dance Style Classification.} 

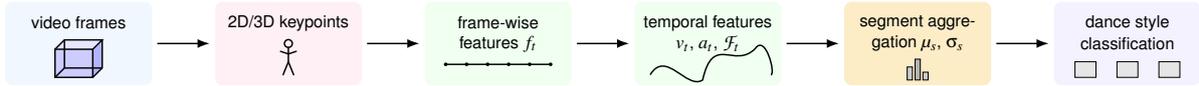
\begin{figure*}[t]
  \centering
  \scalebox{0.69}{
  \begin{tikzpicture}[font=\sffamily,scale=1, every node/.style={transform shape}]

\draw[fill=aliceblue,rounded corners,draw=none] (0,0) rectangle (2.8,1.6);
\node at (1.4,1.2) {video frames};

\begin{scope}[shift={(1.4,0.15)}, scale=0.9]
  \fill[blue!20] (-0.5,0) -- (-0.5,0.6) -- (0.3,0.6) -- (0.3,0) -- cycle; 
  \fill[blue!10] (0.3,0.6) -- (0.5,0.8) -- (-0.3,0.8) -- (-0.5,0.6) -- cycle; 
  \fill[blue!30] (0.3,0) -- (0.5,0.2) -- (0.5,0.8) -- (0.3,0.6) -- cycle; 
  \draw[thick] (-0.5,0) -- (-0.5,0.6) -- (0.3,0.6) -- (0.3,0) -- cycle; 
  \draw[thick] (0.3,0.6) -- (0.5,0.8) -- (-0.3,0.8) -- (-0.5,0.6); 
  \draw[thick] (0.3,0) -- (0.5,0.2) -- (0.5,0.8); 
  \draw[thick] (-0.5,0) -- (-0.3,0.2) -- (0.5,0.2); 
  \draw[thick] (-0.3,0.2) -- (-0.3,0.8); 
\end{scope}

\draw[-Latex,thick] (2.9,0.8) -- (3.9,0.8);

\draw[fill=lavenderblush,rounded corners,draw=none] (4,0) rectangle (6.8,1.6);
\node at (5.4,1.2) {2D/3D keypoints};

\begin{scope}[shift={(5.4,0.35)}, scale=0.5, line cap=round, line join=round, thick]
  \draw (0,0.9) circle [radius=0.15];
  \draw (0,0.75) -- (0,0.2);
  \draw (-0.3,0.6) -- (0,0.5) -- (0.3,0.6);
  \draw (0,0.2) -- (-0.2,-0.3);
  \draw (0,0.2) -- (0.2,-0.3);
\end{scope}

\draw[-Latex,thick] (6.9,0.8) -- (7.9,0.8);

\draw[fill=honeydew,rounded corners,draw=none] (8,0) rectangle (10.8,1.6);
\node at (9.4,1.2) {frame-wise};
\node at (9.4,0.8) {features $f_t$};

\draw[thick] (8.4,0.4) -- (10.4,0.4);
\foreach \x in {8.4,8.8,9.2,9.6,10.0,10.4} 
  \fill[black] (\x,0.4) circle (1pt);

\draw[-Latex,thick] (10.9,0.8) -- (11.9,0.8);

\draw[fill=honeydew,rounded corners,draw=none] (12,0) rectangle (14.8,1.6);
\node at (13.4,1.2) {temporal features};
\node at (13.4,0.8) {$v_t$, $a_t$, $\mathcal{F}_t$};

\draw[thick] (12.3,0.2) .. controls (12.6,0.7) and (13.0,-0.1) .. (13.3,0.1).. controls (13.5,1.0) and (13.9,0.3) .. (14.3,0.7) .. controls (14.5,1.0) and (14.7,0.2) .. (14.6,0.2);

\draw[-Latex,thick] (14.9,0.8) -- (15.9,0.8);

\draw[fill=bananamania,rounded corners,draw=none] (16,0) rectangle (18.8,1.6);
\node at (17.4,1.2) {segment aggre-};
\node at (17.4,0.8) {gation $\mu_s$, $\sigma_s$};

\begin{scope}[shift={(17.4,0.1)}, scale=0.5]
  \draw[fill=gray!40] (-0.4,0) rectangle (-0.2,0.5);
  \draw[fill=gray!60] (-0.1,0) rectangle (0.1,0.8);
  \draw[fill=gray!40] (0.2,0) rectangle (0.4,0.3);
\end{scope}

\draw[-Latex,thick] (18.9,0.8) -- (19.9,0.8);

\draw[fill=magnolia,rounded corners,draw=none] (20,0) rectangle (22.8,1.6);
\node at (21.4,1.2) {dance style};
\node at (21.4,0.8) {classification};

\foreach \x/\y in {20.6/0.3,21.4/0.3,22.2/0.3} 
  \draw[fill=black!10] (\x-0.2,\y-0.15) rectangle (\x+0.2,\y+0.15);

\end{tikzpicture} }
  \caption{Schematic illustration of our dance style classification pipeline. In the first step, body keypoints are extracted from the input videos. In the next step, frame-wise features are extracted and then temporally related. Finally, these temporal features are divided into segments and fed into the classifiers.}
  \label{fig:method}
\end{figure*}

Research into dance motion understanding involves several related but distinct tasks, including dance gesture recognition, which focuses on identifying short, targeted movements \cite{Raptis2011}, and style or genre classification, which aims to distinguish subtle expressive and rhythmic differences between dance movement patterns \cite{Gupta2024}. 
In this paper, we deal with the classification of dance styles and provide an overview of the current state of research in this field. 
Initial approaches in this area are based on handcrafted features extracted by, for example, spatio-temporal interest points or optical flow of the videos, and use these features as input for multi-class support vector machines \cite{Gupta2024}. 
These methods are sensitive to lighting, perspective, and background variations and are therefore limited in their recognition of complex, dynamic dance movements. 
For this reason, recent approaches are based on deep neural networks and can be divided into two methodological approaches. 
On the one hand, convolutional neural networks are trained directly on the video data end-to-end for the classification task \cite{Guo2025,Gupta2024}. Thus, the model independently learns to extract the relevant visual features automatically capturing complex spatio-temporal patterns such as movement dynamics, body postures, or contextual information. 
On the other hand, classification is performed based on pose estimation features, which offer a structured and interpretable representation of movement, as they model posture and dynamics via the position of keypoints. This abstraction significantly reduces data complexity and eliminates interfering visual factors such as background, lighting, or clothing, making the models more robust and generalizable. This approach is therefore particularly suitable for studying movement quality, rhythm, and style characteristics that are independent of physical appearances. 
In \cite{Mifsud2021}, 18 2D body points are extracted per frame and fed directly (without an intermediate step of feature generation) into a Long Short-Term Memory (LSTM) neural network. The authors distinguish between only two different genres of dance in their experiments, ballet and breakdance. 
In comparison, the authors of \cite{Dewan2018} extract 14 2D keypoints and normalize them with respect to the body skeleton and a reference origin, from which LMA-inspired features are derived and passed to an LSTM for classification. 
In \cite{Turab2025}, a more complex architecture is proposed that uses 3D pose estimation, 3D human mesh reconstruction, and floor-aware body modeling to reconstruct the dancer’s body surface, and extract features that are fed into a random forest classifier or a support vector machine. 

The latter two works are most similar to our approach. In difference to \cite{Dewan2018}, we compare the application of 2D and 3D keypoints, do not need a normalization (as we only determine relative features) and utilize more lightweight models for the classification task than LSTM neural networks. 
In comparison to \cite{Turab2025}, we only use 3D human keypoints and no other reconstructions, just a quickly computable set of handcrafted features.
Towards both works, we test several diverse datasets (not just one dance style dataset respectively), and, in particular, we do not only use features based on LMA, but also create frequency-domain motion features based on FFT to identify repetitive movement patterns, dominant frequencies, and rhythmic structures.
%
%
%
\section{\uppercase{Method Description}}
\label{sec:method}
To extract features from the dance videos, the first step is to extract the keypoints of the human body from the individual video frames using motion capture devices or pose estimation networks. 
An overview of our method is shown in Figure~\ref{fig:method}. 
Using extracted keypoints instead of complete RGB video data offers several advantages for classifying dance styles. First, pose-based representations focus directly on the dynamics of the human body and filter out irrelevant background information, lighting variations, and clothing differences that can mislead pixel-based models. Second, the dimensionality of the data is drastically reduced, from high-resolution image sequences to compact skeleton coordinates, enabling more efficient training and inference, especially with limited datasets. 
Let $\mathbf{k}_t^j$ denote the 2D coordinate $(x_t^j,y_t^j)$ or the 3D coordinate $(x_t^j,y_t^j,z_t^j)$ respectively of each joint $j \in \{1,\ldots,N_j\}$ in frame $t$, where $N_j$ is the number of extracted keypoints. In the following, we present features per frame $t$ that reflect each of the four categories of the LMA: body, shape, space and effort. 

\textbf{Body} \hspace{0.5ex} describes which parts of the body move and how they are coordinated with each other, i.e., the organization and connection of movement in the body. 
To quantify joint connectivity, we determine the relative distances between pairs of keypoints.
The hip center is given by
\begin{equation}
  \mathbf{c}_t^\mathit{hip} = \frac{1}{2} \Bigl( \mathbf{k}_t^\mathit{hip\_left} + \mathbf{k}_t^\mathit{hip\_right} \Bigr) \enspace ,
\end{equation}
and the distance of joint $j$ from the hip center in computed by
\begin{equation}
  d_t^j= \lVert \mathbf{k}_t^j - \mathbf{c}_t^\mathit{hip} \rVert_2 \ , \ \forall \ j \in \{ 1,\ldots,N_j \} \enspace .
  \label{eq:dist}
\end{equation}
This representation captures the structural relationships within the body regardless of its absolute position in the image, which eliminates the need for normalization of the coordinates.

\textbf{Shape} \hspace{0.5ex} refers to the spatial change of the body in space, i.e., how the body shape unfolds, expands, or contracts during movement.
To represent the body shape, the distances between the extremities are calculated, meaning between the hands and feet, as well as crosswise
\begin{align}
 e_t^{\mathit{hands}} = & \lVert \mathbf{k}_t^{\mathit{hand\_right}} - \mathbf{k}_t^{\mathit{hand\_left}} \rVert_2 \enspace , \notag\\ 
 e_t^{\mathit{feet}} = & \lVert \mathbf{k}_t^{\mathit{foot\_right}} - \mathbf{k}_t^{\mathit{foot\_left}} \rVert_2 \enspace , \notag\\
 e_t^{\mathit{cross}} = & \tfrac{1}{2} \Bigl(
   \lVert \mathbf{k}_t^{\mathit{hand\_right}} - \mathbf{k}_t^{\mathit{foot\_left}} \rVert_2 \notag\\
   & + \lVert \mathbf{k}_t^{\mathit{hand\_left}}  - \mathbf{k}_t^{\mathit{foot\_right}} \rVert_2
 \Bigr) \enspace .
\end{align}
This representation captures the structural proportions of the body, whether the body is more opened or closed.

\textbf{Space} \hspace{0.5ex} describes the spatial orientation and direction of movement of the body, i.e., how the movement is positioned and oriented in space (e.g., straight, circular, in different planes). 
To determine the alignment of the body in space, the angle is calculated, which describes the rotation of the upper body around the vertical axis $y$. 
We define the main axis of the torso based on the vector from the center of the hips to the center of the chest $\mathbf{o}_t = \mathbf{c}_t^\mathit{chest} - \mathbf{c}_t^\mathit{hip}$. 
To obtain the horizontal body orientation, we project $\mathbf{o}_t$ onto the $xz$-plane, which represents the ground plane in the camera coordinate system. 
The resulting projection $\mathbf{o}_t^{xz} = (o_x,0,o_z)$ is used to compute the yaw angle $\varphi_t$ defined as
\begin{equation}
    \varphi_t = \operatorname{atan2}(o_x, o_z) \enspace ,
\end{equation}
where $o_x$ and $o_z$ denote the horizontal components of the torso vector. 
This angle $\varphi_t \in [-\pi,\pi]$ represents the rotation of the upper body around the vertical axis and thus indicates the facing direction of the dancer relative to the camera.

\textbf{Effort} \hspace{0.5ex} characterizes the quality of movement, i.e., how energy, dynamics, and intention are used (e.g., strong vs.\ gentle, fast vs.\ slow). 
To capture the dynamic properties of motion, we quantify velocity, acceleration, and rhythmic patterns for predefined features, thereby describing the temporal evolution of body movements, such as energy and intensity. 
Our complete set of consists of $N_j+4$ features and is given by
\begin{align}
     f_t \in & \{ d_t^j \big| j \in \{ 1,\ldots,N_j \} \} \notag\\
     & \cup \{e_t^{\mathit{hands}}, e_t^{\mathit{feet}}, e_t^{\mathit{cross}} \} \cup \{ \varphi_t \} \enspace .
\end{align}
For each frame-wise feature $f_t$, we derive the velocity
\begin{equation}
    v_t = \frac{f_t-f_{t-1}}{\Delta t} \ , \ \forall \ t \in \{2,\ldots,T \} \enspace ,
\end{equation}
and the acceleration 
\begin{equation}
    a_t = \frac{v_t-v_{t-1}}{\Delta t} \ , \ \forall \ t \in \{3,\ldots,T \} \enspace ,
\end{equation}
where $\Delta t$ denotes the time difference between two consecutive frames and $T$ the number of frames. 
Note, since the rotation angle $\varphi_t$ is derived using the $\operatorname{atan2}$ function, its range is limited to $[-\pi,\pi]$. As a result, rotations exceeding this limit can lead to discontinuities between consecutive images, resulting in abrupt jumps in the calculated angular velocity. To ensure temporal consistency, the discontinuities are corrected by adding or subtracting multiples of $2\pi$ when the change from frame to frame exceeds the value $\pi$, resulting in a continuous representation. 

In addition, we apply the Fast Fourier Transform to the mean-normalized temporal signals to extract frequency-domain features
\begin{equation}
    \mathcal{F}_t = \operatorname{FFT} ( f_t - \bar f ) \ , \ \forall \ t \in \{1,\ldots,T \} \enspace ,
\end{equation}
where $\bar f = 1/T \sum_{t=1}^T f_t$ represents the mean over the respective feature over time. 
The resulting spectral information captures repetitive or oscillating patterns that are characteristic of dance, such as rhythmic movements of the limbs or periodic turns. 
Combined, these temporal and spectral features provide a compact and expressive representation of movement dynamics and form the basis for distinguishing between dance styles with different energetic and rhythmic profiles. 

Instead of using the frame-wise features directly as input for the classifier, we divide the sequence into $N_s$ segments and calculate the mean $\mu_s$ and standard deviation $\sigma_s$ of the features for each segment $s$, $s=1,\ldots,N_s$, with a length of $T_s = T / N_s$ each. 
The advantages of this approach are that it significantly reduces the dimensionality of the input data and makes the classifier more robust to small variations in movement or outliers. Simultaneously, local movement patterns and recurring sequences are preserved, which can be crucial for distinguishing between dance styles. 
In comparison, aggregation across the entire sequence would result in a loss of this dynamic information, while the use of frame-wise features would generate a very large and challenging input representation. 
In summary, we obtain $N_j+4$ features per frame directly $f_t$, velocities $v_t$, acceleration $a_t$, and FFT $\mathcal{F}_t$, from which we then calculate the mean and standard deviation, thus obtaining $(N_j+4)\cdot 4 \cdot 2$ features per segment.
%
%
%
\section{\uppercase{Experiments}}
\label{sec:exp}
First, we present the experimental setup and then analyze our method in terms of its ability to classify different dance styles.
%
%
%
\subsection{Experimental Setting}
\paragraph{Datasets.} 
The AIST dataset \cite{Tsuchida2019_aist} is a large-scale 3D dance motion database (created using multiple camera angles) covering $10$ different dance genres, including Break, Pop, Lock, Waack (old school), Middle Hip-hop, LA-style Hip-hop, House, Krump, Street Jazz and Ballet Jazz (new school). 
We utilize the diversity of solo dancer videos, with $141$ clips for each dance style, divided into two degrees of complexity, i.e., $120$ basic clips and $21$ advanced clips. 
Each dance choreography comes in four variations, meaning different songs with different speeds. We carefully create non-overlapping train and test splits to ensure that neither choreography nor music is shared between the subsets. 
We use the 2D (front camera only) and 3D body keypoints extracted from the videos in the AIST++ database \cite{Li2021_aist}, with $N_j=17$ keypoints per frame and frame rate of $\Delta t=1/60$ seconds. 

The Motorica Dance Dataset \cite{Alexanderson2023,Perez2022} covers $8$ different styles, some already known from the AIST dataset such as Hip-hop, Krump, Pop, Lock and Jazz, as well as others like Charleston, Tapping and Casual. $121$ dance clips are provided, without distinction between different expert levels, at a frame rate of $120$ fps and with skeleton annotations in the Biovision Hierarchy format. We extract the $N_j=7$ relevant 3D keypoints \emph{hip, right shoulder, left shoulder, right hand, left hand, right foot and left foot} for feature calculation. 

The ImperialDance-Dataset \cite{Zhong2024} consists of $5$ different styles, i.e., Ballet, Hip-hop, Jazz, K-Pop and Urban, spread across $20$ different dance clips, with dancers from all levels (expert, intermediate, and beginner) for each clip. The motion camera records the coordinates of $N_j=21$ joints of the skeleton in 3D space at $100$ fps.
\paragraph{Classification Models.} 
We feed our motion features into various  classification models, i.e., logistic regression with $\ell_{1}$-penalization (LR, \cite{Tibshirani1996}), random forest with $100$ trees (RF), XGBoost with a maximum depth of $6$ and a learning rate of $0.3$ (GB, \cite{Chen_2016}) and shallow neural network containing only a single hidden layer with $500$ neurons with \textit{ReLU} activation function and the Adam \cite{Kingma2014AdamAM} optimizer applied for $300$ iterations (NN).

\paragraph{Evaluation.} 
For evaluation, we adopt a 5-fold cross-validation with 80/20 split strategy to ensure robust and reliable performance estimation. As the primary evaluation metric, we use classification accuracy, which measures the proportion of correctly predicted dance styles relative to all test samples. 

Each dance video is divided into several temporal segments $N_s$, from which features are extracted independently. During training and testing, these segments are treated as individual samples to increase data diversity and capture temporal variations in the performances. To determine the final predicted class of a complete video, we apply a majority voting method to the predictions of the corresponding segments.
We ensure that consistent divisions are maintained during cross-validation, i.e., all segments from the same video are assigned exclusively to either the training or test set to avoid data leakage. 
For the initial experiments, we divide each video clip into $10$ segments of equal length ($N_s=10$) in order to achieve a balance between effort and accuracy. An ablation study is performed later.
%
%
%
\subsection{Numerical Results}
\begin{table*}[t] 
\caption{A performance comparison of the different classifiers and training/test data splits using our complete set of features for the 3D keypoints calculated over 10 segments per sequence for the AIST dataset. ``LR'' refers to logistic regression, ``RF'' to random forest, ``GB'' to gradient boosting, and ``NN'' to neural network. For the different expert levels, ``B'' stands for basic, ``A'' for advanced, and ``M'' for a a combination of the two. Values in percentage.}
\label{tab:aist_3d} 
\centering 
\scalebox{0.73}{ \begin{tabular}{c ccccccccc} \toprule model & \multicolumn{9}{c}{dataset split (train / test)} \\ \cmidrule(r){1-1} \cmidrule(r){2-10} 
& B / B & B / A & B / M & A / B & A / A & A / M & M / B & M / A & M / M \\ \cmidrule(r){2-10} 
LR & $ 86.02 \pm 5.27$ & $ 53.36 \pm 2.64$ & $ 81.22 \pm 4.72$ & $ 49.93 \pm 6.65$ & $ 95.98 \pm 3.37$ & $ 56.63 \pm 5.81$ & $ 89.18 \pm 3.08$ & $ 92.97 \pm 3.93$ & $ 89.72 \pm 3.01$ \\ 
RF & $ 88.42 \pm 3.40$ & $ 60.46 \pm 5.60$ & $ 84.38 \pm 3.32$ & $ 55.49 \pm 6.53$ & $ 97.01 \pm 2.38$ & $ 61.55 \pm 5.80$ & $ 89.40 \pm 3.10$ & $ 93.46 \pm 4.88$ & $ 89.95 \pm 2.93$ \\ 
GB & $ 91.64 \pm 1.55$ & $ \mathbf{62.98} \pm 5.48$ & $ \mathbf{87.52} \pm 1.76$ & $ \mathbf{60.05} \pm 5.18$ & $ 98.04 \pm 2.85$ & $ \mathbf{65.53} \pm 4.91$ & $ \mathbf{93.25} \pm 1.49$ & $ \mathbf{98.99} \pm 1.23$ & $ \mathbf{94.11} \pm 1.27$ \\ 
NN & $ \mathbf{91.74} \pm 2.68$ & $ 39.01 \pm 6.65$ & $ 84.04 \pm 2.32$ & $ 51.23 \pm 10.36$ & $ \mathbf{99.00} \pm 1.23$ & $ 58.24 \pm 8.63$ & $ 93.15 \pm 2.13$ & $ 98.97 \pm 2.05$ & $ 94.01 \pm 1.71$ \\ \bottomrule \end{tabular} } \end{table*} 
\begin{figure}[t]
  \centering
  \includegraphics[width=\linewidth]{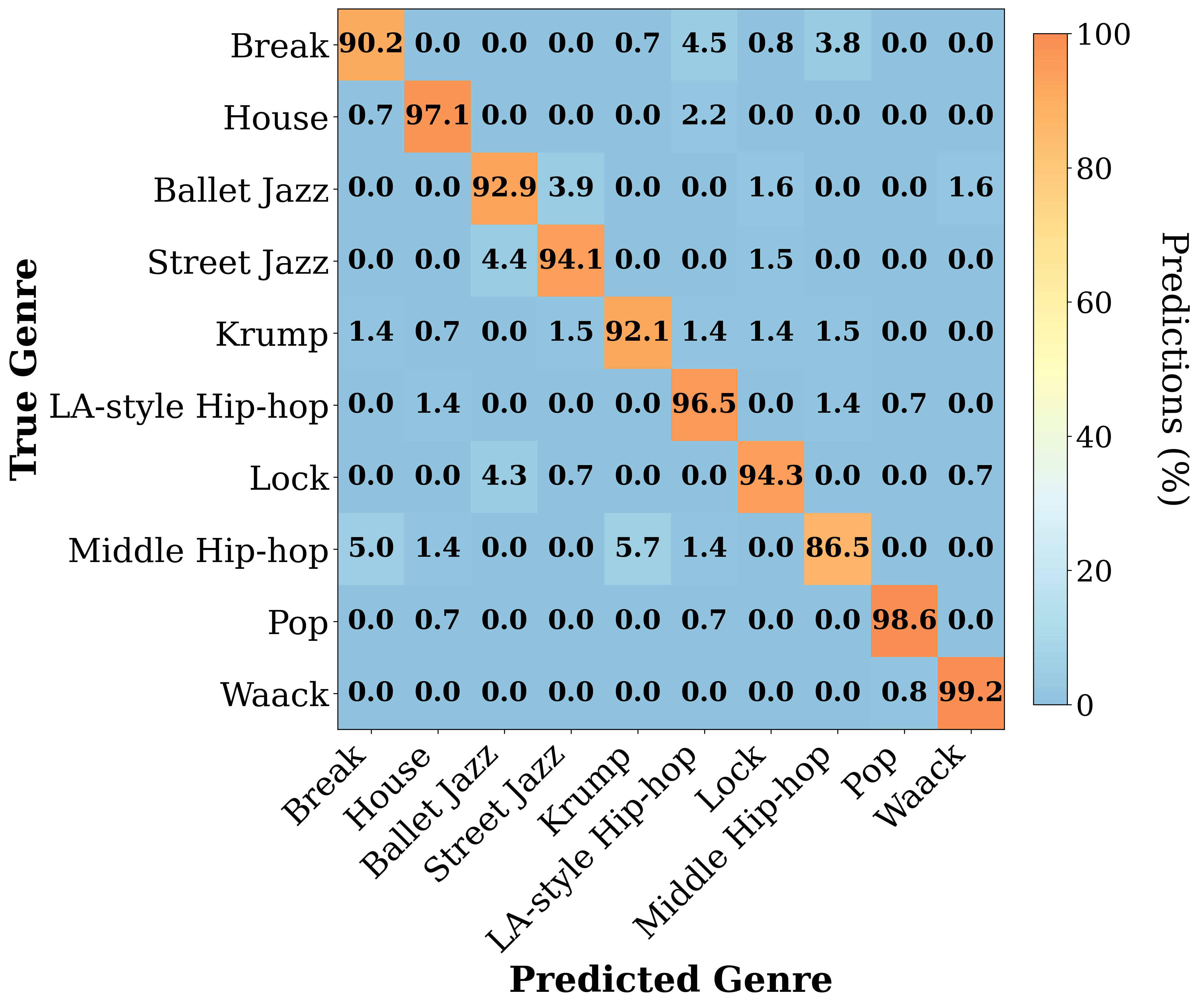}
  \caption{Confusion matrix for the different dance genres of the AIST dataset using gradient boosting and train/test split of mixed/mixed.}
  \label{fig:conf_classes}
\end{figure}
%
%
\begin{table*}[t]
\caption{Comparison with baselines for the AIST dataset. The classifiers are fed with (1) 3D keypoints (without feature calculation), (2) 2D keypoints for feature computation, (3) 3D keypoints for feature computation without FFT features, and (4) the complete feature set using 3D keypoints (ours).}
\label{tab:aist_comparison} 
\centering
\scalebox{0.73}{
\begin{tabular}{lc ccc}
  \toprule 
  method & model & B / B & A / A & M / M \\
  \cmidrule(r){1-1} \cmidrule(r){2-2} \cmidrule(r){3-5}
  3D keypoints $\mathbf{k}_t^j$ & GB & $68.77\pm 0.60$ & $41.00\pm 1.14$ & $70.89\pm 0.54
$ \\
   & NN & $58.70\pm 0.55$ & $26.99\pm 0.60$ & $57.73 \pm 0.38$ \\
  \midrule 
  2D features & GB & $73.83 \pm 4.13$ & $92.92 \pm 3.65$ & $79.67 \pm 4.42$ \\
   & NN & $74.20 \pm 2.63$ & $94.87 \pm 3.72$ & $82.30 \pm 2.04$ \\
  \midrule 
   3D features w/o $\mathcal{F}_t$ & LR & $85.05 \pm 4.43$ & $96.02 \pm 3.29$ & $88.27 \pm 3.12$ \\
   & RF & $89.81 \pm 1.95$ & $94.97 \pm 2.63$ & $91.50 \pm 1.44$ \\
   & GB & $90.99 \pm 1.48$ & $99.01 \pm 1.21$ & $93.97 \pm 1.44$ \\
   & NN & $92.34 \pm 2.49$ & $99.00 \pm 1.23$ & $94.08 \pm 0.64$ \\
  \midrule 
   3D features & LR & $86.02 \pm 5.27$ & $95.98 \pm 3.37$ & $89.72 \pm 3.01$ \\
   & RF & $88.42 \pm 3.40$ & $97.01 \pm 2.38$ & $89.95 \pm 2.93$ \\
   & GB & $91.64 \pm 1.55$ & $98.04 \pm 2.85$ & $94.11 \pm 1.27$ \\
   & NN & $91.74\pm2.68$ & $99.00\pm2.05$ & $94.00\pm1.71$ \\
  \bottomrule
\end{tabular} }
\end{table*}
\paragraph{Dance Style Classification for AIST.} 
Table~\ref{tab:aist_3d} summarizes the quantitative evaluation of our proposed feature representations on the AIST dataset across the different train/test combinations for the \emph{basic} (B), \emph{advanced} (A), and \emph{mixed} (M) subsets. Each result represents the mean~$\pm$~standard deviation of the cross-validation using the complete set of features from the 3D keypoints aggregated over $10$ temporal segments ($N_s = 10$). 
When comparing the variation between the models, clear trends are observed in both, the average performance and the associated standard deviations. Gradient boosting and neural network classifiers achieve the highest overall accuracies up to $99.00\%$ and exhibit the smallest deviations across folds. 
This low dispersion indicates that the models generalize consistently across the five folds and are not overly sensitive to individual split compositions. 
As expected, low variances occur when training and evaluation are performed on the same dataset, and high values are found in particular when training is performed on basic and evaluation on advanced, and vice versa. 
These higher fluctuations indicate increased uncertainty when transferring models between varying levels of expertise, reflecting the pronounced stylistic and dynamic differences between basic and advanced choreographies, which is also directly reflected in the lower accuracies. 
Across all models, the mixed training setup yields the most stable generalization with accuracies greater than $93.25\%$, characterized by both, high mean accuracy and low variability. 
The good performance is explained as the models have seen both expert levels, and thus capture representative stylistic features common to both basic and advanced dances. 

A corresponding confusion matrix for the different dance genres is given in Figure~\ref{fig:conf_classes}. We obtain very high accuracies on the diagonal, i.e, the predicted genres correspond to the true ones, with values between $86.5\%$ and $99.2\%$. 
On the one hand, we observe that some styles clearly differ from others, such as Waack, Pop and House. On the other hand, there is confusion between Middle Hip-hop and Break, as well as Middle Hip-hop and Krump, which indicates the similarity between these dance styles. 
Overall, the results show that combining 3D spatial information with LMA and FFT features achieves very high results in distinguishing between different dance styles.

\begin{table*}[t]
\caption{\emph{Left:} Ablation over number of segments $N_s$ for the AIST dataset with all features based on the 3D keypoints and gradient boosting as classifier. \emph{Right:} Ablation using the two different evaluation methods. ``MV'' describes the majority voting method used in previous experiments. ``TI'' describes that we input the features across the different segments as an additional temporal component into the models and obtain only one prediction per video.}\label{tab:aist_blation} 
\centering
\scalebox{0.73}{
\begin{tabular}{cccc}
  \toprule 
  $N_s$ & B / B & A / A & M / M \\
  \cmidrule(r){1-1} \cmidrule(r){2-4} 
  $1$ & $78.93\pm2.40$ & $79.87\pm8.56$ & $82.23\pm4.05$ \\
  $5$ & $87.26\pm1.57$ & $97.98\pm1.01$ & $90.76\pm0.85$ \\
  $10$ & $91.64\pm1.55$ & $98.04\pm2.85$ & $94.11\pm1.27$ \\
  $15$ & $93.62\pm1.64$ & $98.52\pm 1.96$ & $96.35\pm 0.78$ \\
  $20$ & $93.78 \pm 1.97$ & $99.01 \pm 1.21$ & $96.48 \pm 0.44$ \\
  \bottomrule
\end{tabular} }
\scalebox{0.73}{
\begin{tabular}{cccc}
  \toprule 
  eval & B / B & A / A & M / M \\
  \cmidrule(r){1-1} \cmidrule(r){2-4} 
  MV & $91.64 \pm 1.55$ & $98.04 \pm 2.85$ & $94.11 \pm 1.27$ \\
  TI & $77.23 \pm 6.07$ & $67.70 \pm 11.41$ & $79.90 \pm 4.74$ \\
  \bottomrule
\end{tabular} }
\end{table*}
\paragraph{Comparison with Baselines.} 
We base our approach on similar work and compare our method with baselines. 
Firstly, we compare the use of 2D vs. 3D keypoints for feature extraction. 
Note, the yaw angle $\varphi_t$ features, which describes the torso orientation in space, cannot be calculated in the 2D case and are therefore ignored. 
Secondly, instead of using our defined handcrafted features, we feed all raw body keypoints directly into the classification models to examine how strongly performance depends on feature structuring. 
Finally, we assess the impact of extending the LMA-based features with frequency-domain descriptors obtained from the FFT. 
Note that in comparison to \cite{Turab2025}, which is most similar to our approach, we do not use complex 3D reconstruction, and create other as well as extend the LMA-based features. The authors also conducted tests using the AIST dataset, but not all videos were used, the division into ``basic'' and ``advanced'' was not taken into account, and the redundancy of the same choreographies and music pieces was not considered, making a direct comparison unfeasible. 

\begin{figure}[t]
  \centering
  \includegraphics[width=0.9\linewidth]{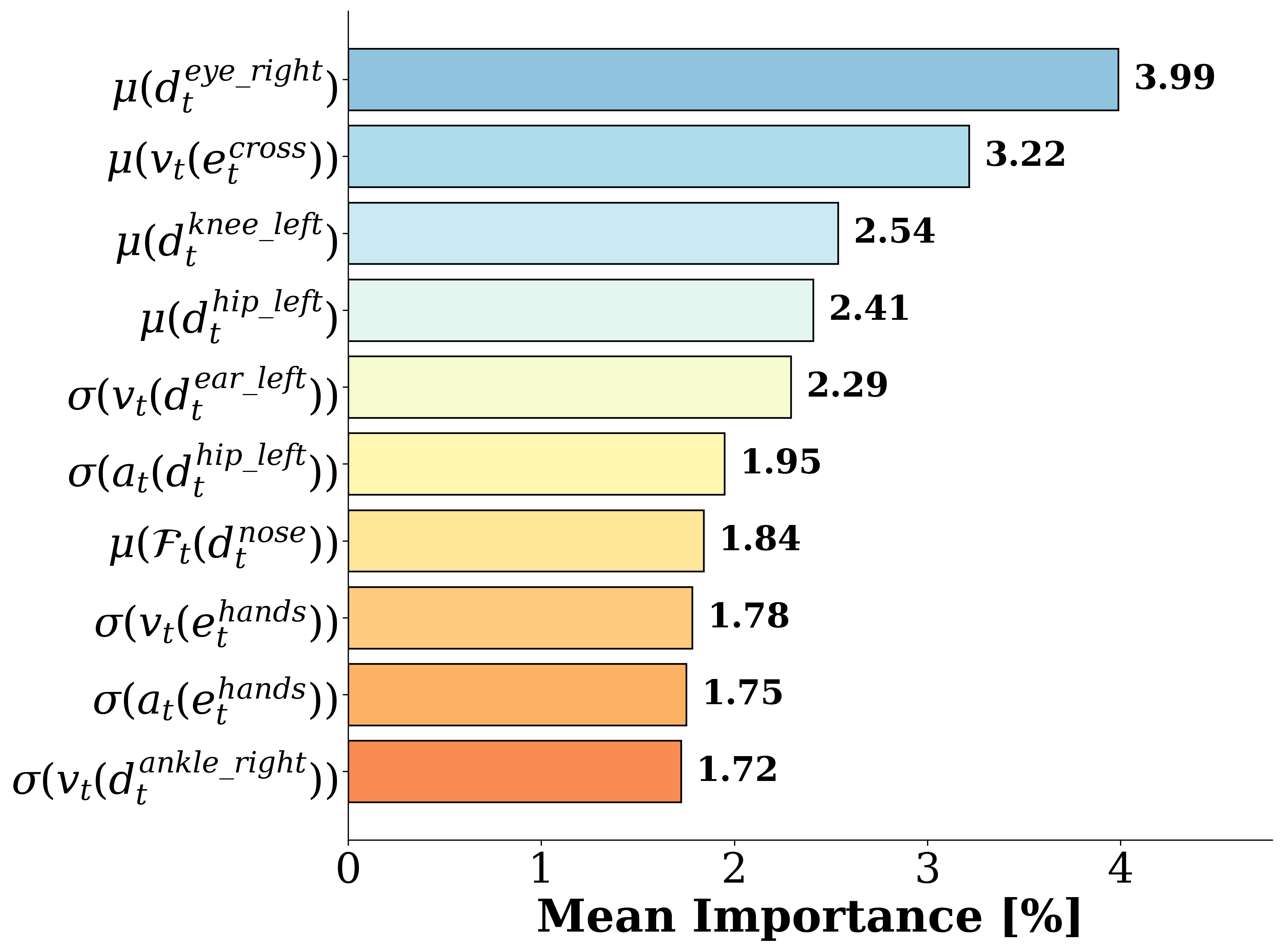}
  \caption{Mean feature importance plot for the AIST dataset using the full feature set from 3D keypoints, gradient boosting as classifier and train/test split of mixed/mixed.}
  \label{fig:xgb_importance}
\end{figure}

Table~\ref{tab:aist_comparison} compares the performance of our proposed feature representation against several baseline configurations. 
The lowest prediction accuracies (regarding M/M) are obtained for the raw 3D keypoints as input, with values of at most $70.89\%$. 
This indicates that the models receive too much information due to the large number of keypoints per frame to form meaningful features for classification, or that they need significantly more complex models. 
Instead of obtaining $(17+4)\cdot 4 \cdot 2 \cdot 10 = 1,\!680$ features when a sequence is divided into $10$ segments, the model obtains $17 \cdot 3 \cdot 2,\!878= 146,\!778$ keypoints as input when the sequence has a length of $48$ seconds.

When features are generated based on 2D keypoints, this leads to significantly lower accuracy and higher variation compared to 3D, which can be explained by the loss of depth and orientation information. 
The accuracies are between $4.13$ and $17.81$ percentage points lower compared to using the 3D information. 
Adding FFT features to the LMA-based features for the 3D keypoints does not improve the neural network performance, as it may be capable of creating similar features itself. 
In comparison, for more lightweight classifiers such as linear regression and gradient boosting, the frequency domain features provide a performance improvement of up to $2.04$ percentage points. 
In Figure~\ref{fig:xgb_importance}, a mean feature important plot for gradients boosting is given, depicting the $10$ features with the greatest influence on the prediction. 
It is a combination of distance features, but also speeds and accelerations, and in particular, an FFT feature has a high influence. 
Overall, the comparison with the baselines shows that our set of features effectively captures the differences between dance genres and achieves the highest classification accuracies.

\paragraph{Ablation Studies.} 
In the following, we will perform two ablations. First, we will study the influence of the number of segments in which the dance sequences are embedded, and second, we will examine how we treat the segments in the evaluation.

In Table~\ref{tab:aist_blation} (left), the ablation over the number of segments is given using gradient boosting as classifier and the complete feature set as input. We varied $N_s$ from $1$ to $20$ while keeping all other settings fixed. 
The case $N_s=1$ corresponds to the features being averaged over the entire clip resulting in temporal correlations being dropped, which is also reflected in the performance, achieving accuracies between only $78.93\%$ and $82.23\%$ percent. 
As the number of segments increases, accuracy improves across all train/test splits. The reason is that the models receive more motion information per clip and are therefore more capable of identifying differences in style. 
Performance stagnates at a segment count of $15$, which means that more segments provide little additional useful information. 
Note that doubling the number of segments, e.g., from $10$ to $20$, also results in twice as many features being fed into the classifiers, which is why we limited ourselves to $N_s=10$ for the other experiments. With this value, we achieve comparatively high results with significantly fewer features than with more splits of the sequences. 

We further compared two training/evaluation strategies. On the one hand, the majority voting (MV) method, which we have used in previous experiments, treats the individual segments of each sequence as separate samples and applies a majority decision method to the predictions of the corresponding segments to determine the final predicted class of a complete video. 
On the other hand, for temporal integration (TI), we input the features across the different segments as an additional temporal component into the models and obtain only one prediction per video. 
The results are shown in Table~\ref{tab:aist_blation} (right) using gradient boosting and a number of segments of $N_s=10$. 
Majority voting generally achieves higher accuracies of at least $14.21$ percentage points compared to temporal integration. 
The majority voting approach ensures that the independent treatment of segments prior to aggregation provides a more consistent decision-making mechanism, and that short-term fluctuations or misclassifications within individual segments have only a limited influence on the overall decision.

\begin{table}[t] 
\centering
\caption{A performance comparison of the different classifiers and segment counts $N_s$ for the Motorica Dance Dataset and the ImperialDance-Dataset.}
\scalebox{0.73}{
\begin{tabular}{cc ccc}
\toprule
dataset & model & \multicolumn{3}{c}{segment count $N_s$} \\
\cmidrule(r){1-1} \cmidrule(r){2-2} \cmidrule(r){3-5}
& & $5$ & $10$ & $15$ \\
\cmidrule(r){3-5}
Motorica & LR & $89.48\pm 7.78$ & $89.52\pm 10.32$ & $85.71\pm8.91$\\
Dance    & RF & $84.71\pm 6.98$ & $85.71\pm 9.52$ & $83.71\pm6.27$ \\
Dataset  & GB & $89.52\pm6.21$ & $91.38\pm6.17$ & $88.52\pm5.36$\\
         & NN & $87.57\pm6.32$ & $88.57\pm 9.28$ & $92.38\pm7.22$\\
\midrule
Imperial   & LR & $95.63\pm 0.59$ & $94.22 \pm 0.49$ & $92.44\pm 0.61$ \\
Dance-     & RF & $98.33\pm 0.29$ & $97.83 \pm 0.52$ & $97.89\pm 0.48$ \\
Dataset    & GB & $99.25 \pm 0.14$ & $99.62 \pm 0.10$ & $99.62 \pm 0.06$ \\
           & NN & $99.07 \pm 0.07$ & $99.41 \pm 0.27$ & $99.54 \pm 0.18$ \\
\bottomrule
\end{tabular}}
\label{tab:imperial_motorica}
\end{table}
\paragraph{Results for More Datasets.} 
The classification results for the Motorica Dance Dataset and the ImperialDance-Dataset are given in Table~\ref{tab:imperial_motorica} using our complete set of features and the majority voting procedure. 
Note, due to the small number of choreographies for the latter dataset (only $20$ pieces), we do not differentiate between the expert levels in the results. 
As with the AIST dataset, gradient boosting and the neural network outperform the other two methods. 
We observe that for these two datasets, the number of segments into which we split each sequence has little effect on performance. 
We achieve very high accuracies of up to $92.38\%$ for the Motorica Dance Dataset (with the reduced keypoint set) and up to $99.62\%$ for the ImperialDance-Dataset.

%
%
%
\section{\uppercase{Conclusion}}
\label{sec:conc}
In this paper, we have presented a lightweight approach for dance style classification based on interpretable features derived from poses. Instead of processing raw RGB data, we relied on 3D keypoints obtained from motion capture devices or human pose estimation networks, from which we extracted features that (i) are inspired by Laban movement analysis and (ii) encode rhythmic and periodic aspects using Fast Fourier Transform. These features captured spatial relationships, body rotations, and dynamic aspects such as velocity, acceleration, and rhythmic movement patterns. By aggregating these feature descriptions across temporal slices, we obtained compact and robust representations that balance motion detail and computational efficiency. 

Our experiments show that this feature-based representation provides a meaningful basis for distinguishing dance genres while ensuring low computational cost and high interpretability. Unlike deep end-to-end architectures, the proposed method enables explicit analysis of motion qualities, which is of great value for applications in dance education, motion analysis, and performance studies.

In future work, we plan to extend this framework by incorporating temporal modeling techniques such as recurrent or attention-based networks to better capture long-term dependencies in motion sequences. Furthermore, the integration of multi-modal information, such as music context, could further improve classification accuracy, and enable a deeper understanding of the relationship between rhythm, style, and movement expression.


\bibliographystyle{apalike}
{\small
\bibliography{example}}

@article{Alexanderson2023,
  title={Listen, Denoise, Action! Audio-Driven Motion Synthesis with Diffusion Models},
  author={Alexanderson, Simon and Nagy, Rajmund and Beskow, Jonas and Henter, Gustav Eje},
  year={2023},
  issue_date={August 2023},
  publisher={ACM},
  volume={42},
  number={4},
  doi={10.1145/3592458},
  journal={ACM Trans. Graph.},
  articleno={44},
  numpages={20},
  pages={44:1--44:20}
}

@ARTICLE{Brigham1967,
  author={Brigham, E. O. and Morrow, R. E.},
  journal={IEEE Spectrum}, 
  title={The fast Fourier transform}, 
  year={1967},
  volume={4},
  number={12},
  pages={63-70},
  doi={10.1109/MSPEC.1967.5217220}
}

@inproceedings{Dewan2018_lma,
author = {Swati Dewan and Shubham Agarwal and Navjyoti Singh},
title = {{Laban movement analysis to classify emotions from motion}},
volume = {10696},
booktitle = {Tenth International Conference on Machine Vision (ICMV 2017)},
editor = {Antanas Verikas and Petia Radeva and Dmitry Nikolaev and Jianhong Zhou},
organization = {International Society for Optics and Photonics},
publisher = {SPIE},
pages = {106962Q},
year = {2018},
doi = {10.1117/12.2309451},
URL = {https://doi.org/10.1117/12.2309451}
}

@INPROCEEDINGS{Dewan2018,
  author={Dewan, Swati and Agarwal, Shubham and Singh, Navjyoti},
  booktitle={24th International Conference on Pattern Recognition (ICPR)}, 
  title={Spatio-Temporal Laban Features for Dance Style Recognition}, 
  year={2018},
  volume={},
  number={},
  pages={2911-2916},
  doi={10.1109/ICPR.2018.8545251}
}

@article{Guo2025,
author = {Guo, Na and Yang, Ahong and Wang, Yan and Dastbaravardeh, Elaheh},
year = {2025},
month = {04},
pages = {},
title = {Fine-Grained Dance Style Classification Using an Optimized Hybrid Convolutional Neural Network Architecture for Video Processing Over Multimedia Networks},
volume = {2025},
journal = {International Journal of Intelligent Systems},
doi = {10.1155/int/6434673}
}

@article{Gupta2024,
title = {Indian dance classification using machine learning techniques: A survey},
journal = {Entertainment Computing},
volume = {50},
pages = {100639},
year = {2024},
issn = {1875-9521},
doi = {https://doi.org/10.1016/j.entcom.2024.100639},
url = {https://www.sciencedirect.com/science/article/pii/S1875952124000077},
author = {Sharish Gupta and Sarbjeet Singh},
}

@article{Jobanputra2019,
title = {Human Activity Recognition: A Survey},
journal = {Procedia Computer Science},
volume = {155},
pages = {698-703},
year = {2019},
note = {The 16th International Conference on Mobile Systems and Pervasive Computing (MobiSPC 2019),The 14th International Conference on Future Networks and Communications (FNC-2019),The 9th International Conference on Sustainable Energy Information Technology},
issn = {1877-0509},
doi = {https://doi.org/10.1016/j.procs.2019.08.100},
url = {https://www.sciencedirect.com/science/article/pii/S1877050919310166},
author = {Charmi Jobanputra and Jatna Bavishi and Nishant Doshi},
}

@misc{Li2021_aist,
      title={Learn to Dance with AIST++: Music Conditioned 3D Dance Generation}, 
      author={Ruilong Li and Shan Yang and David A. Ross and Angjoo Kanazawa},
      year={2021},
      eprint={2101.08779},
      archivePrefix={arXiv},
      primaryClass={cs.CV}
}

@INPROCEEDINGS{Mifsud2021,
  author={Mifsud, Yanika and Inguanez, Frankie},
  booktitle={2021 IEEE 11th International Conference on Consumer Electronics (ICCE-Berlin)}, 
  title={Dance Style Classification by LSTM RNN}, 
  year={2021},
  volume={},
  number={},
  pages={1-6},
  doi={10.1109/ICCE-Berlin53567.2021.9720031}
}

@inproceedings{Perez2025,
author = {Perez-Martinez, Roberto and Casas-Ortiz, Alberto and Santos, Olga C.},
title = {MoRTELaban: a Neurosymbolic Framework for Motion Representation and Analysis based on Labanotation and Laban Movement Analysis},
year = {2025},
isbn = {9798400713996},
publisher = {Association for Computing Machinery},
address = {New York, NY, USA},
url = {https://doi.org/10.1145/3708319.3734180},
doi = {10.1145/3708319.3734180},
booktitle = {Adjunct Proceedings of the 33rd ACM Conference on User Modeling, Adaptation and Personalization},
pages = {353–359},
numpages = {7},
location = {},
series = {UMAP Adjunct '25}
}

@article{Perez2022,
  author={Valle-P{\'e}rez, Guillermo and Henter, Gustav Eje and Beskow, Jonas and Holzapfel, Andre and Oudeyer, Pierre-Yves and Alexanderson, Simon},
  title={Transflower: Probabilistic Autoregressive Dance Generation with Multimodal Attention},
  year={2021},
  issue_date={December 2021},
  publisher={ACM},
  volume={40},
  number={6},
  doi={10.1145/3478513.3480570},
  journal={ACM Trans. Graph.},
  articleno={195},
  numpages={14},
  pages={195:1--195:14}
}

@inproceedings{Raptis2011,
author = {Raptis, Michalis and Kirovski, Darko and Hoppe, Hugues},
title = {Real-time classification of dance gestures from skeleton animation},
year = {2011},
isbn = {9781450309233},
publisher = {Association for Computing Machinery},
address = {New York, NY, USA},
url = {https://doi.org/10.1145/2019406.2019426},
doi = {10.1145/2019406.2019426},
booktitle = {Proceedings of the 2011 ACM SIGGRAPH/Eurographics Symposium on Computer Animation},
pages = {147–156},
numpages = {10},
location = {Vancouver, British Columbia, Canada},
series = {SCA '11}
}

@article{Tibshirani1996,
author = {Tibshirani, Robert},
year = {1996},
journal = {Journal of the Royal Statistical Society: Series B},
pages = {267--288},
title = {Regression Shrinkage and Selection via the Lasso},
volume = {58},
}

@inproceedings{Tsuchida2019_aist,
           author = {Shuhei Tsuchida and Satoru Fukayama and Masahiro Hamasaki and Masataka Goto}, 
           title = {AIST Dance Video Database: Multi-genre, Multi-dancer, and Multi-camera Database for Dance Information Processing}, 
           booktitle = {Proceedings of the 20th International Society for Music Information Retrieval Conference, {ISMIR} 2019},
           address = {Delft, Netherlands}, 
           year = 2019, 
           month = nov 
}

@misc{Turab2025,
      title={Dance Style Recognition Using Laban Movement Analysis}, 
      author={Muhammad Turab and Philippe Colantoni and Damien Muselet and Alain Tremeau},
      year={2025},
      eprint={2504.21166},
      archivePrefix={arXiv},
      primaryClass={cs.CV},
      url={https://arxiv.org/abs/2504.21166}, 
}

@article{Zheng2023,
author = {Zheng, Ce and Wu, Wenhan and Chen, Chen and Yang, Taojiannan and Zhu, Sijie and Shen, Ju and Kehtarnavaz, Nasser and Shah, Mubarak},
title = {Deep Learning-based Human Pose Estimation: A Survey},
year = {2023},
issue_date = {January 2024},
publisher = {Association for Computing Machinery},
address = {New York, NY, USA},
volume = {56},
number = {1},
issn = {0360-0300},
url = {https://doi.org/10.1145/3603618},
doi = {10.1145/3603618},
journal = {ACM Comput. Surv.},
month = aug,
articleno = {11},
numpages = {37},
}

@article{Zhong2024, 
title={DanceMVP: Self-Supervised Learning for Multi-Task Primitive-Based Dance Performance Assessment via Transformer Text Prompting}, 
volume={38}, url={https://ojs.aaai.org/index.php/AAAI/article/view/28893}, 
DOI={10.1609/aaai.v38i9.28893}, 
number={9}, journal={Proceedings of the AAAI Conference on Artificial Intelligence}, 
author={Zhong, Yun and Demiris, Yiannis}, 
year={2024}, 
month={Mar.}, 
pages={10270-10278}
}

@inproceedings{Chen_2016, series={KDD ’16},
   title={XGBoost: A Scalable Tree Boosting System},
   url={http://dx.doi.org/10.1145/2939672.2939785},
   DOI={10.1145/2939672.2939785},
   booktitle={Proceedings of the 22nd ACM SIGKDD International Conference on Knowledge Discovery and Data Mining},
   publisher={ACM},
   author={Chen, Tianqi and Guestrin, Carlos},
   year={2016},
   month=aug, pages={785–794},
   collection={KDD ’16} }

@article{Kingma2014AdamAM,
  title={Adam: A Method for Stochastic Optimization},
  author={Diederik P. Kingma and Jimmy Ba},
  journal={CoRR},
  year={2014},
  volume={abs/1412.6980},
  url={https://api.semanticscholar.org/CorpusID:6628106}
}



\end{document}